\documentclass[11pt]{article}

\usepackage[final]{acl}

\usepackage{times}
\usepackage{latexsym}
\usepackage[T1]{fontenc}
\usepackage[utf8]{inputenc}
\usepackage{microtype}
\usepackage{inconsolata}

\usepackage{amsmath}
\usepackage{booktabs}
\usepackage{multirow}
\usepackage{makecell}
\usepackage{graphicx}
\usepackage{tikz}
\usepackage[most]{tcolorbox}
\usetikzlibrary{positioning,arrows.meta,fit,backgrounds}

\newlength{\demopanelheight}
\newcommand{\setdemopanelheight}{%
  \setlength{\demopanelheight}{\dimexpr0.5\textwidth*1000/1596\relax}%
}
\newcommand{\demopanelimgqa}[1]{%
  \includegraphics[width=\linewidth,height=\demopanelheight,keepaspectratio]{#1}%
}
\newcommand{\demopanelimgrel}[1]{%
  \includegraphics[width=\linewidth,height=\demopanelheight,keepaspectratio]{#1}%
}

\usepackage{listings}
\usepackage{xcolor}
\newtcolorbox{promptbox}{
  enhanced,
  breakable,
  colback=gray!8,
  colframe=gray!30,
  boxrule=0.3pt,
  arc=0pt,
  left=3pt,
  right=3pt,
  top=3pt,
  bottom=3pt,
  before skip=4pt,
  after skip=4pt,
  fontupper=\ttfamily\scriptsize,
  before upper={\raggedright\setlength{\parindent}{0pt}\sloppy\obeylines}
}

\usepackage{url}
\title{VidEngram: Agentic Long-Term Video Understanding\\via Structured Hierarchical Memory}

\author{
  Zinuo Cheng$^{1}$\quad Yueqian Lin$^{1}$\quad Zhongxi Wang$^{1}$\quad Hai ``Helen'' Li$^{1}$\quad Yiran Chen$^{1}$ \\
  $^{1}$Duke University \\
  \texttt{\{zinuo.cheng, yiran.chen\}@duke.edu}
}

\begin{document}
\maketitle

\begin{abstract}
Long-video question answering is wasteful and lossy when a multimodal model must
re-encode the same video for every query: context limits force sparse sampling or
truncation, and no reusable representation remains for later questions. We
present \textbf{VidEngram}, an agentic video QA system that builds a persistent
hierarchical memory once at ingest and answers later questions from that memory.
VidEngram captions video and audio jointly with Qwen2.5-Omni, consolidates
segment memories into episode and entity layers, and uses a text-only agentic
planner with specialized retrieval tools to produce grounded answers. The web
interface makes this process inspectable: users can watch memory formation,
inspect retrieval traces, verify cited evidence in the video player, and explore
entity relationships. On the full OmniVideoBench, VidEngram with a GPT-5-mini planner outperforms a strong single-pass Qwen2.5-Omni baseline (37.0\% vs.\
32.3\%), with gains concentrated on longer videos where single-pass coverage
becomes sparse. Because perception is paid once, subsequent queries use low-cost
text-only planning.
\end{abstract}

\section{Introduction}
\label{sec:intro}

\begin{figure}[t]
\centering
\includegraphics[width=\columnwidth]{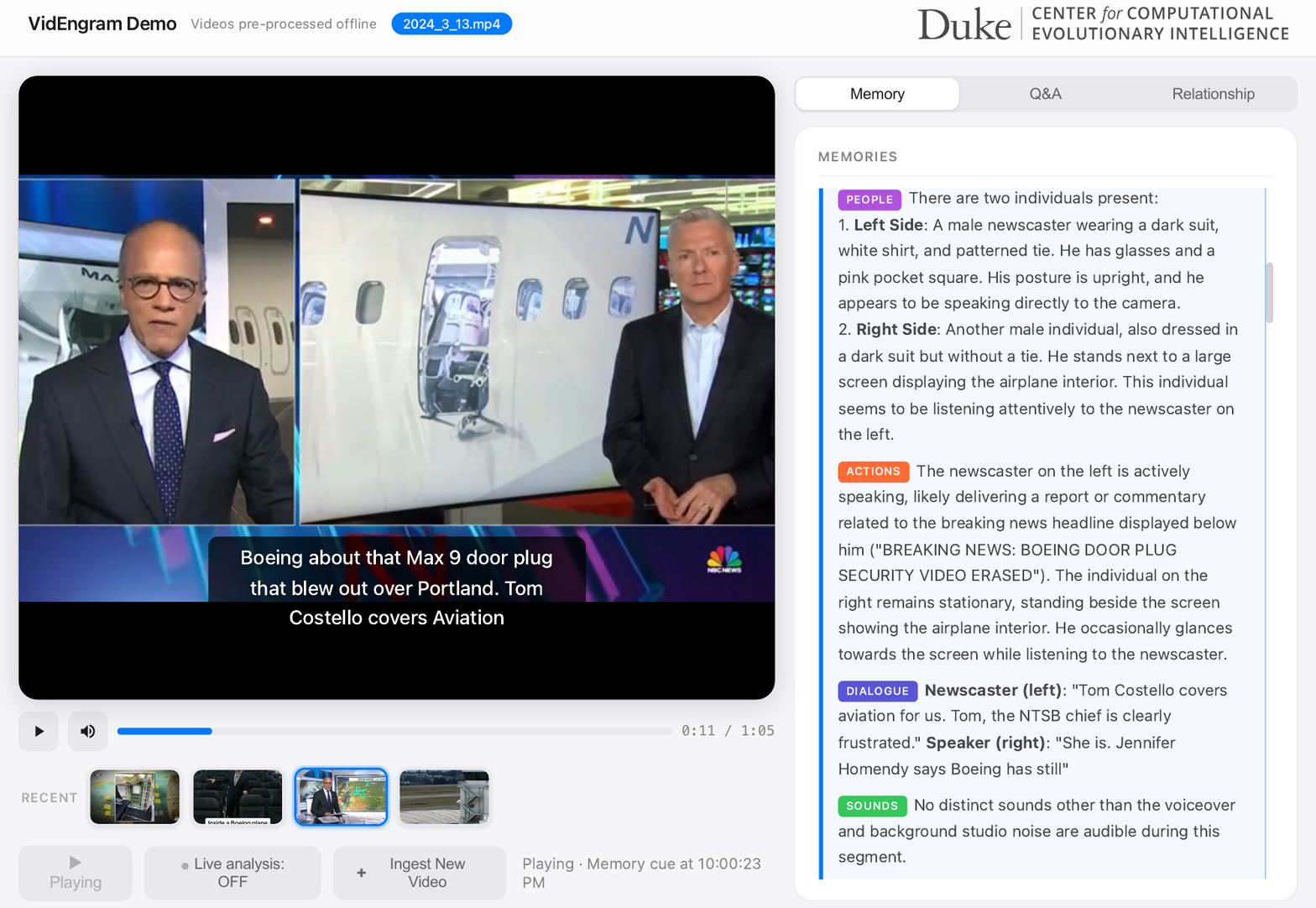}
\caption{Memory formation view. The user watches the source video while the
eleven-field structured-caption feed is populated in real time, making the
stored memory inspectable before later questions are asked.}
\label{fig:teaser}
\end{figure}

Long-video question answering is a repeated-query problem: the same lecture,
documentary, or meeting may be queried many times, but a multimodal model
typically re-encodes the video for every question and compresses it into a
bounded context window. This makes single-pass QA expensive and lossy. Sparse
frame sampling breaks temporal continuity~\cite{videollama2023}, flat RAG
retrieves isolated caption snippets without narrative structure~\cite{vrag2025},
and even long-context models~\cite{gemini15} provide no reusable memory.
The bottleneck is therefore representational, since users need an inspectable
memory that can be reused across questions.

We present \textbf{VidEngram} (Figure~\ref{fig:teaser}), which treats video
understanding as memory formation and retrieval. Rather than compressing a video
into one context window per query, VidEngram encodes it once into a structured,
indexed memory; at query time, a text-only agentic planner selects retrieval tools over that
memory and grounds answers in retrieved evidence. In the web interface,
users can watch memory formation, inspect retrieval traces, jump to cited video
evidence, and explore entity relationships.
Empirically, this separation improves long-video QA without rebuilding memory.
On OmniVideoBench, VidEngram with a GPT-5-mini planner improves overall accuracy
from 32.3\% to 37.0\%, with gains concentrated on medium, long, and ultra-long
videos, at lower query latency than the single-pass baseline.

\paragraph{Contributions.}
\begin{enumerate}
\item \textbf{Reusable video memory.}
  VidEngram converts video and audio into structured segment memories and a
segment\,\textrightarrow\,episode\,\textrightarrow\,entity hierarchy for
reusable retrieval.

\item \textbf{Inspectable agentic retrieval.}
  A ReAct loop~\cite{yao2023react} selects among memory and clip-verification
  tools, exposing retrieval traces and evidence for user inspection.

\item \textbf{Decoupling perception from reasoning.}
  The omni-model runs once at ingest. Each query then uses a swappable text-only
  planner over memory, so planner quality can improve without re-encoding video.
\end{enumerate}

\begin{figure*}[t]
\centering
\includegraphics[width=\textwidth]{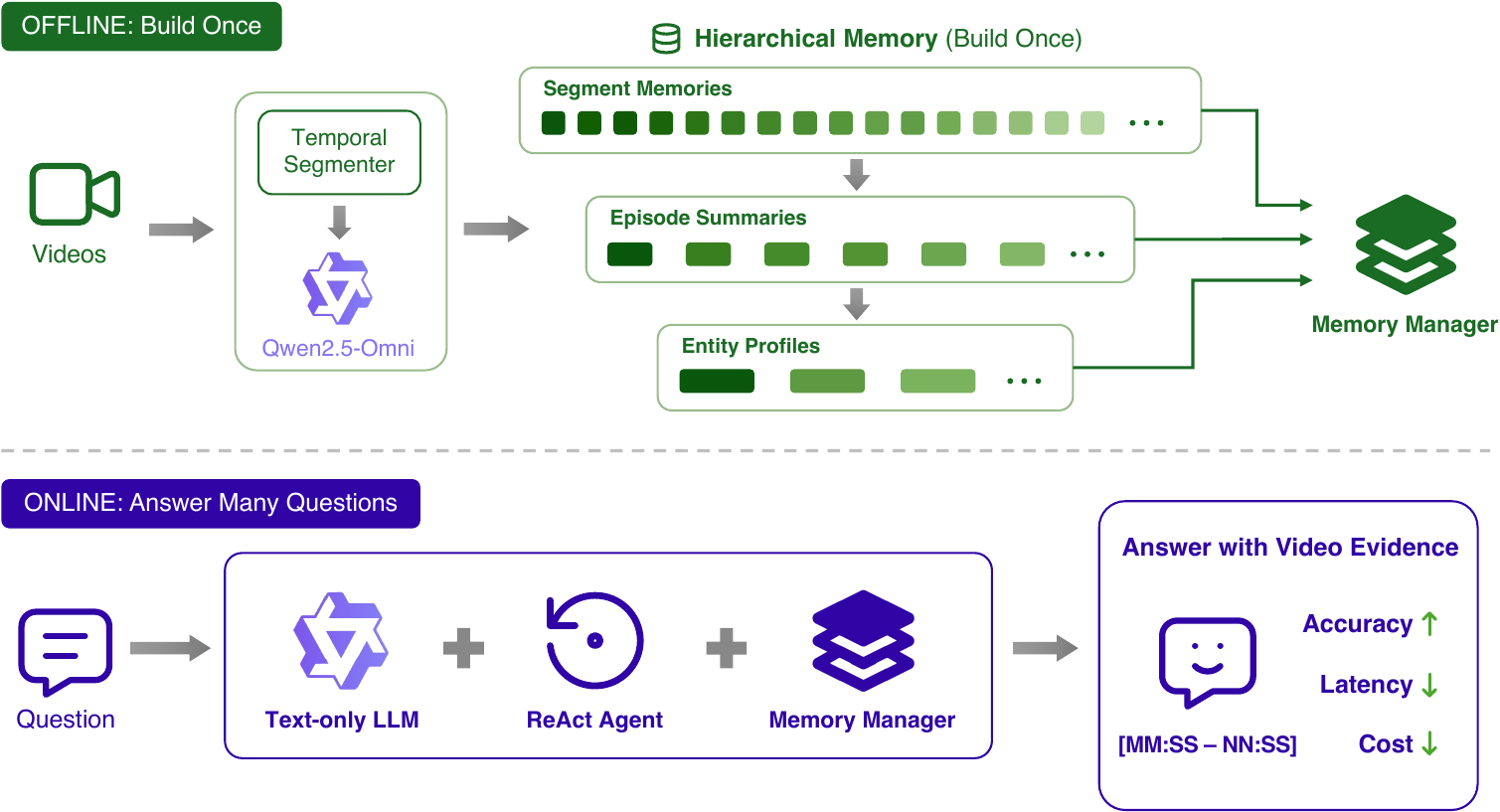}
\caption{VidEngram architecture. \textbf{Offline (build once):} each video is
segmented, captioned by Qwen2.5-Omni into eleven structured fields, and
consolidated into a three-layer memory
(segment\,\textrightarrow\,episode\,\textrightarrow\,entity) in the memory manager.
\textbf{Online (answer many):} a ReAct agent answers each question from memory
through specialized retrieval tools and timestamp citations, so users can inspect
how an answer was produced without re-encoding the full video.}
\label{fig:architecture}
\end{figure*}

\section{Related Work}
\label{sec:related}

\paragraph{Video understanding and hierarchical memory.}
General-purpose vision-language models~\cite{clip2021,flamingo2022,blip2_2023}
and video extensions~\cite{videollama2023,videochatgpt2024,videollava2024,qwen2vl2024,internvideo2_2024}
perform well on short clips, while long-context variants~\cite{longva2024,longvu2024}
extend the input window but still degrade with duration. Agentic and memory-based
systems address this pressure differently: VideoAgent~\cite{videoagent2024} adds
tool use without a persistent backend, VideoARM~\cite{videoarm2026} builds
hierarchical memory at query time through coarse-to-fine backtracking, and
HippoMM~\cite{hippomm2025} introduces an episodic audiovisual hierarchy.
VidEngram instead consolidates a persistent segment/episode/entity store once at
ingest, serves it through keyword and vector indices with cross-video scope, and
uses a single Qwen2.5-Omni pass rather than separate visual and speech encoders.
Related systems build compressed or hierarchical memories~\cite{videolucy2025,memvid2025,videotree2025,moviechat2024,malmm2024}
or retrieve over video~\cite{goldfish2024,llovi2024}, but do not combine a
symbolic episode/entity store, persistent indexing, and agentic timestamped
retrieval.

\paragraph{Retrieval-augmented generation and agentic reasoning.}
Retrieval-augmented generation (RAG)~\cite{rag2020} with dense
retrievers~\cite{dpr2020} has also been adapted to video QA. Video-RAG~\cite{videorag2025}
injects visually aligned auxiliary text into an LVLM in a single pass; VRAG~\cite{vrag2025},
RAVU~\cite{ravu2025}, and VideoRAG~\cite{videoraggraph2025} add retrieval or
graph reasoning over long videos. These systems remain largely single-pass over
flat or graph stores, without LLM-guided narrative consolidation, full-timeline
entity resolution, or an inspectable agentic retrieval loop. VidEngram instead
applies ReAct~\cite{yao2023react} tool use~\cite{toolformer2023} to persistent
structured video memory.

\section{Method}
\label{sec:method}

\subsection{System Overview}
\label{sec:overview}

VidEngram separates video QA into offline memory formation and online memory
retrieval (Figure~\ref{fig:architecture}). During ingest, the system segments a
video, captions each segment with Qwen2.5-Omni, consolidates captions into
segment, episode, and entity layers, and writes these records to a persistent
memory manager~\cite{evermemos2026}. At query time, a text-only agentic planner
searches this memory through specialized tools and returns an answer grounded in
retrieved timestamped evidence.

We use the memory manager only as a persistent storage and hybrid-retrieval
substrate; VidEngram constructs the video-specific memories, controls the
retrieval tools, and handles timestamp-grounded answer presentation. The web interface exposes the same pipeline through Memory, Q\&A, and
Relationship panels (Section~\ref{sec:demo}), while a command-line interface supports batch ingest, scripted evaluation, and
reproducible benchmark queries.

\subsection{Temporal Segmentation}
\label{sec:segmenter}

Temporal segmentation defines the unit of memory formation. VidEngram uses
Whisper~\cite{whisper2023} ASR to segment videos at sentence boundaries. Across our benchmark videos,
this yields approximately 7.7 segments per minute, with an average segment length
of 7--8 seconds. The short segment length is deliberate: it keeps captioning
local enough for precise timestamp citations, while later episode consolidation
restores broader narrative context for retrieval.

\subsection{Unified Audiovisual Encoding}
\label{sec:captioner}

Each segment is encoded as a structured audiovisual memory using
Qwen2.5-Omni-7B~\cite{qwen25omni2025}. Unlike systems such as
HippoMM~\cite{hippomm2025} that run separate visual and speech encoders followed
by fusion, Qwen2.5-Omni processes frames and audio in a single forward pass. This
lets the captioner describe visual events, speech, non-speech sounds, and their
temporal alignment without requiring a separate audiovisual fusion or diarization
module.

Captions use an eleven-field template: \textsc{Scene}, \textsc{People},
\textsc{Actions}, \textsc{Dialogue}, \textsc{Sounds}, \textsc{AV-Sync},
\textsc{Text}, \textsc{Objects}, \textsc{Counts}, \textsc{Emotion}, and
\textsc{Temporal}. We choose these fields to record low-level, directly
observable evidence before memory consolidation, rather than asking the
captioner to infer higher-level topics, causes, or relationships. Those
abstractions are derived later through episode summarization and entity
resolution.

These fields are stored as text rather than collapsed into a single free-form
caption. This preserves different evidence types for later retrieval: exact
dialogue can be found by keyword search, people and appearances can feed entity
profiles, and visual events can be consolidated into episode-level memories.

\subsection{Three-Layer Memory Consolidation}
\label{sec:consolidator}

Caption consolidation turns local segment descriptions into a searchable video
memory. VidEngram stores three memory layers independently: \textbf{Segment}
memories preserve the eleven-field caption and ASR transcript for each short
clip; \textbf{Episode} memories summarize contiguous segments into coherent
narrative units; and \textbf{Entity profiles} merge repeated mentions into
canonical people or objects with aliases, roles, and appearance timelines.

\paragraph{Stage 1: Segment deduplication.}

Adjacent captions above a Jaccard word-set threshold (default 0.7) are merged into
one segment. This reduces redundant near-identical memories in static or
slow-moving scenes while preserving timestamp coverage for downstream retrieval.

\paragraph{Stage 2: Episode grouping.}

Contiguous segments are grouped into episodes, each covering a coherent topic or
scene with a start and end time. After each new segment, an LLM decides whether to
extend the current episode or open a new one, subject to a maximum segment count,
and summarizes each episode into a 3--5 sentence narrative for retrieval and
inspection.

\paragraph{Stage 3: Entity resolution.}
Entity profiles are built in two steps: the LLM first extracts episode-level
mentions with names, roles, and time ranges, then merges these mentions into
canonical profiles with aliases, appearance timelines, and holistic descriptions.
This provides an explicit entity layer for retrieval and relationship inspection,
rather than relying only on a flat list of captions.

The resulting segment, episode, and entity records are written to the memory
manager, tagged by layer, and indexed for hybrid retrieval (Section~\ref{sec:overview}).
A representative 60-minute video produces 483 segment memories, 79 episode
summaries, and 14 entity profiles. Segment memories preserve local evidence and
timestamps, episode memories reduce retrieval fragmentation for scene-level
questions, and entity profiles collect repeated appearances or mentions across
the full timeline.

\subsection{Streaming Ingest Architecture}
\label{sec:streaming}

The ingest pipeline is designed to make memory formation visible during the demo.
VidEngram first runs ASR to obtain sentence boundaries and transcript hints, then
submits the resulting segments to Qwen2.5-Omni for captioning in parallel.
Completed captions are immediately written and indexed as segment memories, so
users can watch the memory store grow rather than waiting for a batch output.
Episode and entity consolidation run incrementally over this stream. The write
rate is bounded mainly by Omni-model parallelism, such as served workers or API
concurrency. On a 60-minute video, the first segment memory is ready in
$\sim$8\,s, versus $\sim$16\,min if all captions must finish before memory is
written.

\subsection{ReAct Orchestrator}
\label{sec:react}

\begin{table}[t]
\centering
\scriptsize
\setlength{\tabcolsep}{2.5pt}
\begin{tabular}{@{}lll@{}}
\toprule
\textbf{Tool} & \textbf{Retrieval} & \textbf{Best for} \\
\midrule
\texttt{search\_episodes} & Hybrid (BM25+vec+RRF) & Factual / visual queries \\
\texttt{search\_speech} & BM25 keyword & Exact word / quote lookup \\
\texttt{search\_profiles} & Hybrid over entity & Who-is / appearance \\
\texttt{search\_deep} & LLM-guided multi-hop & Complex reasoning \\
\texttt{look\_at\_clip} & Qwen2.5-Omni re-anal. & Audio-visual verification \\
\texttt{get\_timeline} & Chronological listing & Time-range summaries \\
\bottomrule
\end{tabular}
\caption{The six tools available to the VidEngram ReAct agent. Hybrid
tools fuse BM25 and vector retrieval via reciprocal rank
fusion; \texttt{look\_at\_clip} verifies a short clip when
memory is insufficient.}
\label{tab:tools}
\end{table}

\begin{figure}[t]
\centering
\includegraphics[width=0.82\columnwidth]{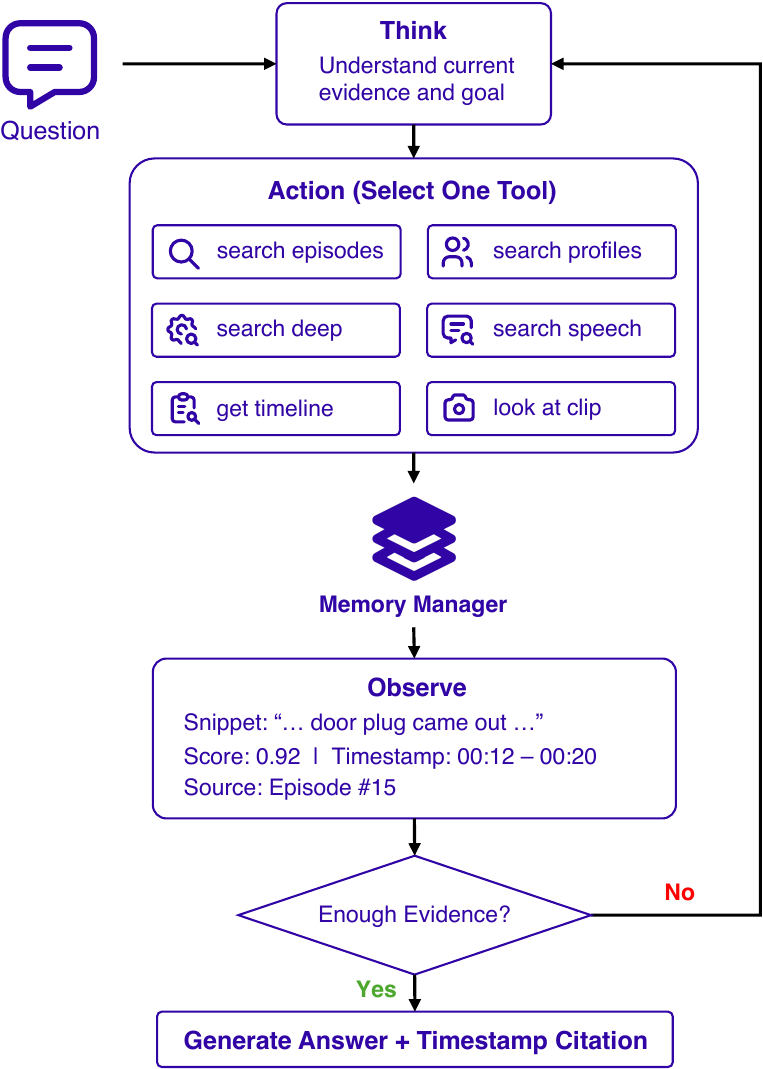}
\caption{ReAct retrieval workflow. The planner selects among six tools; the memory manager returns
evidence, and the agent iterates until it has enough support to answer with a
timestamp citation.}
\label{fig:react-loop}
\end{figure}

At query time, VidEngram answers from memory with a text-only planning LLM in a
ReAct loop (Figure~\ref{fig:react-loop}). Given the question, dialogue context,
video metadata, and tool specification, the planner first issues a retrieval
action, then answers only after observing sufficient retrieved evidence.

\textsc{Think and Action.}
At each iteration, the planner identifies the missing evidence and selects an
appropriate retrieval tool. Table~\ref{tab:tools} lists the available tools and
their intended use cases. The tools retrieve from pre-built segment, episode, and
entity memories. The only exception is \texttt{look\_at\_clip}, which calls
Qwen2.5-Omni only on the final fallback iteration, after memory retrieval has
failed to provide sufficient evidence. In that case, memory retrieval first
narrows the target time span, so Qwen2.5-Omni re-examines only a short candidate
clip rather than the full video.

\textsc{Observe.}
Each tool returns ranked snippets with metadata and canonical timestamps.
Observations are appended to the planner context, where scores rank evidence and
trigger fallbacks but are not treated as calibrated confidence.

\textsc{Decide and Answer.}
The planner then determines whether the collected evidence is sufficient. If
not, it continues the loop with another retrieval action; otherwise, it
generates an answer grounded in the collected evidence. Final answers retain
only citations traceable to segment metadata, so users can inspect the cited
video span. The demo configuration allows up to five iterations before answering from
the strongest observations.

\begin{figure*}[!t]
\centering
\setdemopanelheight
\begin{minipage}[t]{0.47\textwidth}
\centering
\demopanelimgqa{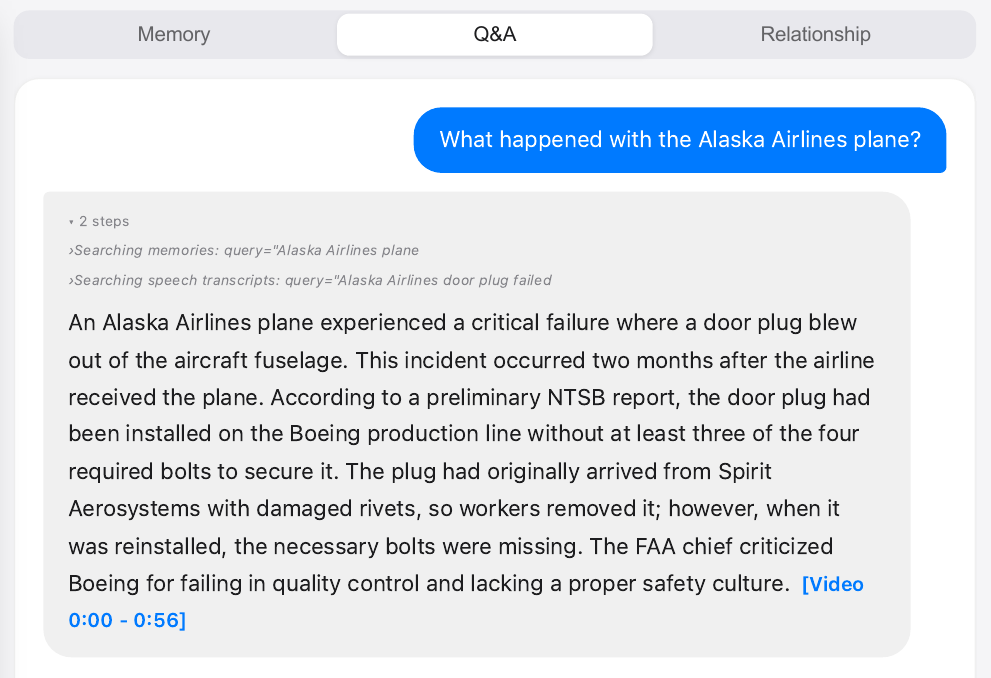}
\par\vspace{1pt}{\small (a) Q\&A case study\par}
\end{minipage}\hspace{0.03\textwidth}%
\begin{minipage}[t]{0.5\textwidth}
\centering
\demopanelimgrel{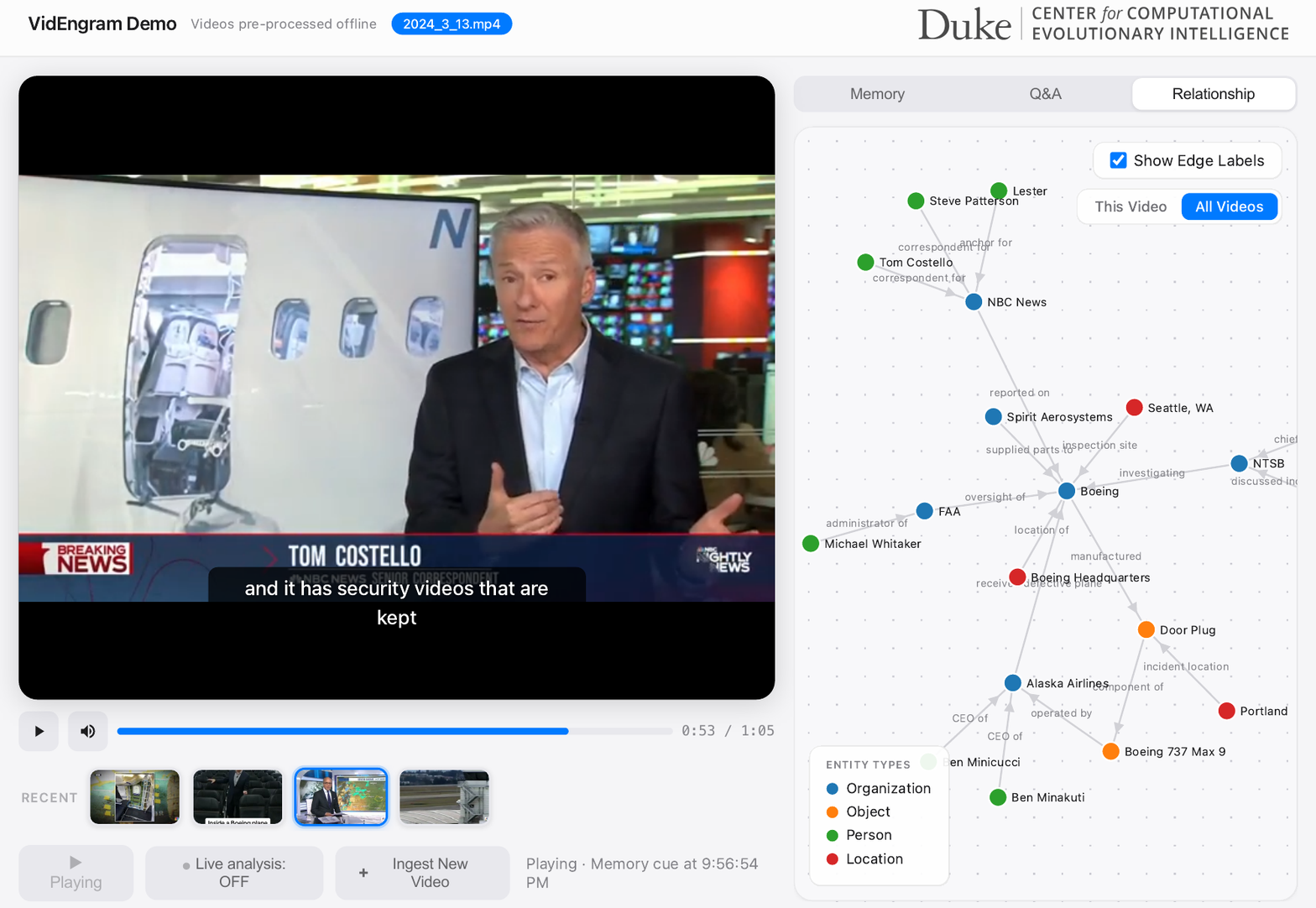}
\par\vspace{1pt}{\small (b) Relationship view\par}
\end{minipage}
\caption{Demonstration views. \textbf{(a)} In the Q\&A case study, a user asks
what happened with the Alaska Airlines plane. VidEngram first searches video
memories for the event, then searches speech transcripts for the more specific
door-plug failure, and returns a grounded answer whose citation can be clicked to
verify the evidence in the video player. \textbf{(b)} The Relationship view
visualizes the entity layer for the same case, linking the entity nodes
with labeled relations; selecting a node opens its profile and source scenes.}
\label{fig:demo-panels}
\end{figure*}

\subsection{Cross-Video Analysis}
\label{sec:crossvideo}

The memory manager's group-id filtering lets the agent search all ingested videos
simultaneously without additional infrastructure. In ``all videos'' mode,
timestamp citations are suppressed for cross-video results, since memories from
different videos do not share a common timeline. The same interface also supports
collection-level relationship analysis by merging entity profiles across the
ingested videos in one graph view.

\section{Demonstration}
\label{sec:demo}

VidEngram is shown as a browser-based demo
(Figures~\ref{fig:teaser} and~\ref{fig:demo-panels}) organized into Memory,
Q\&A, and Relationship panels. The demo video shows memory records, retrieval
tool choices, returned evidence, and cited playback spans.

\paragraph{Memory.}
The Memory panel makes ingestion inspectable. It shows the eleven-field
structured-caption feed beside time-synced playback (Figure~\ref{fig:teaser}).
In the demo video, this feed fills as streaming ingest runs
(Section~\ref{sec:streaming}), and the same interface can show precomputed
records for clips whose memories have already been built.

\paragraph{Q\&A.}
The Q\&A panel exposes the retrieval trace for each answer. In the case study in
Figure~\ref{fig:demo-panels}(a), the question ``What happened with the Alaska
Airlines plane?'' first triggers a broad episode-memory search and then a
targeted speech search for the door-plug failure. The answer combines the two
observations and cites \texttt{[Video 0:00 - 0:56]}, allowing the user to verify
the evidence in the player. This example illustrates why the agent uses multiple
retrieval modes: the first search identifies the event context, while the speech
search recovers the more specific phrase that supports the final answer.

\paragraph{Relationship.}
The Relationship panel exposes the entity layer. It renders extracted entities as
an interactive graph, as shown in Figure~\ref{fig:demo-panels}(b). In the case study,
the graph connects Boeing to the door plug, Alaska Airlines, Boeing Headquarters, FAA, Spirit
AeroSystems, NBC News, Seattle, and NTSB, while labeling the relationships among
these entities. Clicking a node opens its description and appearance timeline,
linking the entity back to source scenes for inspection in the demo. The same
view can switch between a single video and merged cross-video memory, with
cross-video timestamps suppressed when no common timeline exists.

\paragraph{Demo video and code.}
The demo video shows preprocessed clips together with their stored memories,
retrieval traces, Q\&A responses, cited playback spans, and relationship graph.
It also shows the same interface during streaming ingest, where new structured
captions are written into memory as they are produced. A demo
video\footnote{\url{https://youtu.be/hoZQ7Fje5Zw}} and source
code\footnote{\url{https://github.com/No00el/VidEngram}} are available.

\section{Evaluation}
\label{sec:eval}

The evaluation tests whether reusable memory remains robust as videos outgrow
single-pass perception, and whether planner quality affects accuracy without
rebuilding memory.

\begin{table*}[t]
\centering
\begin{tabular}{ll rrrr r @{\qquad} r}
\toprule
 & & \multicolumn{5}{c}{\textbf{Accuracy (\%) by duration tier}} & \textbf{Latency} \\
\cmidrule(lr){3-7}
\textbf{Method} & \textbf{Planning LLM} & \textbf{Short} & \textbf{Med.} & \textbf{Long} & \textbf{Ultra} & \textbf{Overall} & \textbf{med.\,(s)} \\
\midrule
\makecell[l]{Direct-Omni\\ \footnotesize(Qwen2.5-Omni-7B)} & ---  & \textbf{48.5} & 29.9 & 26.0 & 30.7 & 32.3 & 7.0 \\
\midrule
VidEngram & Qwen2.5-7B  & 36.4 & 29.9 & 30.8 & 28.8 & 30.9 & 9.5 \\
VidEngram & GPT-5-mini  & 45.5 & \textbf{37.5} & \textbf{33.5} & 34.1 & 37.0 & \textbf{6.4} \\
VidEngram & GPT-5       & 45.5 & 35.8 & \textbf{33.5} & \textbf{36.7} & \textbf{37.1} & 16.5 \\
\bottomrule
\end{tabular}
\caption{VidEngram on OmniVideoBench (1000 questions, 628 videos). Direct-Omni
  answers in one pass over up to 256 frames plus extracted audio; VidEngram captions
  once with the same Qwen2.5-Omni-7B model and varies only the text planner.
  GPT-5-mini reaches 37.0\% (vs.\ 32.3\%; McNemar $p\,{=}\,0.012$), winning on
  medium, long, and ultra-long videos.}
\label{tab:eval}
\end{table*}

\subsection{Benchmark and Protocol}
\label{sec:eval-setup}

We evaluate on the full OmniVideoBench~\cite{omnivideobench2025}: 628 videos and
1000 multiple-choice questions that couple audio and visual evidence. Scoring is
exact letter match (A/B/C/D), with no LLM judge. Direct-Omni uses Qwen2.5-Omni-7B to answer each question in one pass over up to 256
uniformly sampled frames plus extracted audio, subject to the model processor's
audio limit. This is a strong single-pass recipe under our context and GPU
budget. Both systems answer identical questions, so we use paired McNemar tests
and report relative gains rather than leaderboard rank.
For VidEngram, the Qwen2.5-Omni captioner is held fixed across all rows; only the
text planner changes. This isolates the effect of planner choice at query time:
the same stored memory can be reused with a local Qwen2.5 planner, GPT-5-mini, or
GPT-5 without rebuilding the video index.

\subsection{Results}
\label{sec:eval-results}

VidEngram with a GPT-5-mini planner answers \textbf{37.0\%} of the 1000 questions
correctly, against \textbf{32.3\%} for a strong single-pass Direct-Omni baseline
(256 frames plus extracted audio), a statistically significant $+$4.7-point
margin (paired McNemar $p\,{=}\,0.012$). The advantage appears once a video
exceeds what a single pass can cover
(Table~\ref{tab:eval}): VidEngram trails on short clips ($-$3.0), where one dense
pass already sees and hears the whole clip, but leads on every longer tier
($+$7.6 medium, $+$7.5 long, $+$3.4 ultra-long). A fixed 256-frame budget becomes
sparser as duration grows, while VidEngram answers from segment captions stored
roughly every eight seconds plus the transcript.
This pattern matches the intended use case: memory becomes useful as the same
frame budget must cover longer timelines, where brief visual or spoken events
are more likely to be missed by a capped single pass.

\paragraph{The planning LLM is the key lever.}
The planning LLM is a separate accuracy lever because it can be upgraded without
rerunning the omni-model. Holding the stored memory and tool set fixed, VidEngram scores
30.9\% with Qwen2.5-7B and 37.0\% with GPT-5-mini. The gain saturates beyond
GPT-5-mini: GPT-5 adds only $+$0.1 overall, so GPT-5-mini is the operating point.

\paragraph{Latency is insensitive to video duration.}
Each query retrieves from pre-built memory instead of re-encoding the video, so
VidEngram's median query latency stays near 6.4\,s with the GPT-5-mini planner
and changes little across duration tiers. The capped Direct-Omni baseline also
keeps latency bounded, but only by fixing the input budget; as videos grow
longer, the same cap makes visual and audio coverage increasingly sparse.
Appendix~\ref{app:latency} reports the latency breakdown by duration tier.

\paragraph{Cost is amortized across queries.}
VidEngram pays ingest once and reuses memory with low-cost text-only planning;
re-encoding baselines pay video-processing cost per query. This favors VidEngram
at high query volume. Appendix~\ref{app:cost} gives the breakdown.

\section{Conclusion}
\label{sec:conclusion}

VidEngram demonstrates long-video QA as reusable, inspectable memory. A single
omni-model ingest pass builds a persistent segment--episode--entity hierarchy,
while later questions are answered by a text-only agentic planner over specialized
memory tools. In the demo, users see memory records form during ingest, inspect
the retrieval path behind each answer, and examine entity relationships grounded
in video evidence. On OmniVideoBench, the design is most useful beyond short
clips, where capped single-pass perception becomes sparse, and planner upgrades
improve accuracy without rebuilding memory. These results support VidEngram's separation of one-time video perception from
reusable memory-based reasoning.

\section*{Limitations}
Our accuracy numbers are system-level results, and we do not isolate the
contribution of individual design choices such as the structured caption
template, the memory layers, or the retrieval tools, so the claim that
hierarchical memory preserves information better than single-pass encoding is
supported only indirectly rather than by a same-system ablation. Because each
video is perceived only once at ingest, overall accuracy ultimately reflects the
quality of the omni-model captions. Our
evidence is limited in scope as well, since we report on a single benchmark and
evaluate cross-video reasoning only qualitatively.

\section*{Ethical Considerations}
VidEngram builds persistent, cross-video entity profiles with appearance
timelines, which on identifiable people raise privacy, consent, and
data-retention concerns and enable re-identification across unrelated footage. A
related risk is that these profiles and the answers built on them come from
imperfect models, so the system can misattribute actions, relations, or
identities to a named person, which is more harmful when it is tied to a real
individual than to an anonymous clip. We intend VidEngram for public, archival,
or educational content used with consent, not for surveillance.

\bibliography{refs}

\appendix


\section{Deployment Details}
\label{app:deploy}
The full production
pipeline runs on an eight-GPU server with NVIDIA Quadro
RTX 6000 24GB GPUs. Qwen2.5-Omni-7B is served with vLLM for segment captioning
and bounded clip verification, while text-only models handle planning, embedding,
and reranking. A FastAPI backend writes VidEngram's segment, episode, and entity
records to the memory manager, which provides persistent document, keyword,
vector, and cache services. The demo video shows a lightweight presentation path using preprocessed clips,
where stored memory, Q\&A, relationship views, and agent traces can be inspected
without running the full captioning stack.

\paragraph{Model access.}
The Qwen2.5-Omni-7B captioner, the Qwen2.5-7B planner, and the embedding and
reranking models are self-hosted with vLLM from fixed open weights, whereas the
GPT-5-mini and GPT-5 planners use a commercial API (queried in
June 2026) and may not reproduce exactly as that API changes. All planners answer
the same 1000 questions with the same tools and iteration budget, so only the
planner differs across rows.


\section{Cost Breakdown}
\label{app:cost}

\paragraph{Cost methodology.}
We log every model call and compute cost from measured token usage and published
unit prices; Whisper is billed by processed audio minutes. Latencies in
Table~\ref{tab:eval} are end-to-end wall-clock times from the same tracker.

VidEngram's cost advantage appears along the query dimension. Native 720p video
at $\sim$500--1000 tokens/s would make a one-hour input $\sim$3.6M tokens
(far exceeding the 32K context limit), so full-video QA requires severe
downsampling. The capped Direct-Omni baseline therefore saturates at about
\$0.009/query, while across seven clips (5\,s--30\,min) VidEngram's ingest cost
grows with duration and per-query usage stays at $\approx$8K text tokens. Thus
$C_{\text{total}} = C_{\text{ingest}} + n \cdot c_q$, with
$c_q \approx \$4{\times}10^{-4}$/query: ingest dominates, and each further
question adds only a small text-planning cost, so VidEngram breaks even within
a few queries for clips up to one minute and within $\sim$65 queries even at
30\,min (Table~\ref{tab:cumcost7b}).

Table~\ref{tab:cumcost7b} uses Qwen2.5-7B planning to show a deployable low-cost
setting. GPT-5-mini and GPT-5 are also text-only planners, so switching to them
increases only the small $c_q$ term rather than reprocessing video per query.

\begin{table}[t]
\centering
\footnotesize
\setlength{\tabcolsep}{4pt}
\resizebox{\columnwidth}{!}{%
\begin{tabular}{l rr rr rr}
\toprule
 & \multicolumn{2}{c}{\textbf{1 query}} & \multicolumn{2}{c}{\textbf{10 queries}} & \multicolumn{2}{c}{\textbf{100 queries}} \\
\cmidrule(lr){2-3}\cmidrule(lr){4-5}\cmidrule(lr){6-7}
\textbf{Dur.} & \textbf{VG} & \textbf{BL} & \textbf{VG} & \textbf{BL} & \textbf{VG} & \textbf{BL} \\
\midrule
5\,s              & \$0.002 & \$0.001 & \$0.006 & \$0.010 & \$0.042 & \$0.098 \\
10\,s             & \$0.002 & \$0.002 & \$0.006 & \$0.020 & \$0.042 & \$0.196 \\
30\,s             & \$0.007 & \$0.006 & \$0.010 & \$0.059 & \$0.046 & \$0.588 \\
1\,min            & \$0.016 & \$0.009 & \$0.019 & \$0.092 & \$0.055 & \$0.917 \\
5\,min            & \$0.078 & \$0.009 & \$0.081 & \$0.092 & \$0.117 & \$0.917 \\
10\,min           & \$0.172 & \$0.009 & \$0.176 & \$0.092 & \$0.212 & \$0.917 \\
30\,min           & \$0.574 & \$0.009 & \$0.577 & \$0.092 & \$0.613 & \$0.917 \\
\bottomrule
\end{tabular}}
\caption{Cumulative measured cost (USD) over $n$ queries: VidEngram (VG,
Qwen2.5-7B planning) vs.\ a Direct-Omni baseline (BL). VG pays a one-time ingest
cost plus a small per-query planning cost; BL re-pays the capped video cost on
every query.}
\label{tab:cumcost7b}
\end{table}

\section{Query Latency by Duration}
\label{app:latency}
Table~\ref{tab:latency-by-duration} breaks down median query latency by
OmniVideoBench duration tier over the same 1000 questions as
Table~\ref{tab:eval}. Since VidEngram answers from pre-built memory, planner
choice is the main query-time factor; GPT-5-mini and GPT-5 latencies include
commercial-API overhead.

\begin{table}[h]
\centering
\small
\setlength{\tabcolsep}{3.5pt}
\resizebox{\columnwidth}{!}{%
\begin{tabular}{l rrrrr}
\toprule
\textbf{Method} & \textbf{Short} & \textbf{Med.} & \textbf{Long} & \textbf{Ultra} & \textbf{Overall} \\
\midrule
Direct-Omni & 6.34 & 6.56 & 7.06 & 7.16 & 6.99 \\
VidEngram + Qwen2.5-7B & 8.65 & 9.67 & 9.73 & 9.54 & 9.49 \\
VidEngram + GPT-5-mini & 5.81 & 6.69 & 6.55 & 6.21 & 6.43 \\
VidEngram + GPT-5 & 11.98 & 15.91 & 18.24 & 19.00 & 16.51 \\
\bottomrule
\end{tabular}}
\caption{Median query latency (seconds) by OmniVideoBench duration tier.}
\label{tab:latency-by-duration}
\end{table}

\section{Captioning Prompt}
\label{app:prompt}
The eleven-field structured caption (Section~\ref{sec:captioner}) is produced by the
boxed system prompt below, sent to Qwen2.5-Omni-7B, followed in
deployment by a worked example segment (omitted for space) and a user message
instructing the model to emit all eleven fields at length.

\begin{promptbox}
You are a precise, EXHAUSTIVE video analyst. For each video segment output EXACTLY these labeled fields, each on its own line. Record EVERY detail -- exact counts and quantities, colors, on-screen text, spatial relationships, object attributes, and fine-grained actions. Downstream questions are detailed and counting-heavy and will depend on the specifics you capture here, so err strongly on the side of MORE detail. NEVER summarize -- ENUMERATE the specifics instead (e.g. "red cape, blue mask, yellow emblem"; "3 books and 2 red cups"; "upbeat piano with drums").

SCENE: [Physical setting, location, lighting, camera angle and movement, background elements. Be exhaustive.]
PEOPLE: [Every visible person: identity/name if known, clothing and colors, position, expression, posture. State how many there are, and note relative attributes (tallest/shortest, leftmost/rightmost, oldest/youngest, nearest/farthest) so later questions can disambiguate which person.]
ACTIONS: [What each person does, all interactions and movements, described step by step in order.]
DIALOGUE: [When an ASR transcript is provided, use it verbatim; attribute each utterance to the visually identifiable speaker (e.g. "John:", "Woman in red:"), else "Speaker A:", "Speaker B:". Transcribe yourself if no ASR is provided. "None" if the segment is silent.]
SOUNDS: [Every non-speech sound in detail: animal/object sounds, music (genre, instruments, tempo, mood), sound effects, ambient noise. Count occurrences.]
AV\_SYNC: [Audio-visual co-occurrence: for every notable sound, speech line, or music cue, state what is visually happening at that exact moment and who is involved (identified by distinguishing attributes such as tallest or red shirt), linking each audio event to the simultaneous visual with "->". "None" only if the segment is fully silent.]
TEXT: [ALL visible on-screen text, signs, titles, subtitles -- transcribe verbatim. "None" if absent.]
OBJECTS: [ALL notable objects, tools, props: their colors, quantities, positions, and any text on them.]
COUNTS: [Explicit counts of anything countable: number of people, repeated objects, occurrences of an event. Always give numbers when possible.]
EMOTION: [Emotional tone with specific descriptors and nuance.]
TEMPORAL: [Sequence of events within this segment in order: what happens first, what changes, how it ends.]
\end{promptbox}

\end{document}